\documentclass{article}

\usepackage[square,sort,comma,numbers]{natbib}
\usepackage{amsmath}
\usepackage{graphicx}
\usepackage{arxiv}
\usepackage{setspace}
\usepackage[utf8]{inputenc} 
\usepackage[T1]{fontenc}    
\usepackage{hyperref}       
\usepackage{url}            
\usepackage{booktabs}       
\usepackage{amsfonts}       
\usepackage{nicefrac}       
\usepackage{microtype}      
\usepackage{lipsum}
\usepackage{mathtools}
\usepackage{epigraph}
\usepackage{csquotes}
\usepackage{xcolor}
\usepackage{fancybox}

\date{}

\title{Do Large GPT Models Discover Moral Dimensions in Language Representations? A Topological Study Of Sentence Embeddings.}

\author{
    \bf Stephen Fitz \\
    \texttt{mail@stephenfitz.net} \\
}

\begin{document}

\maketitle
\onehalfspacing

\begin{abstract}
  Since the release of ChatGPT, and subsequently GPT-4, there has been an increased interest in AI safety and alignment research. As Large Language Models are deployed within Artificial Intelligence systems, that are increasingly integrated with human society, it becomes more important than ever to study their internal structures. Higher level abilities of LLMs such as GPT-3.5 emerge in large part due to informative language representations they induce from raw text data during pre-training on trillions of words. These embeddings exist in vector spaces of several thousand dimensions, and their processing involves mapping between multiple vector spaces, with total number of parameters on the order of trillions. Furthermore, these language representations are induced by gradient optimization, resulting in a black box system that is hard to interpret. In this paper, we take a look at the topological structure of neuronal activity in the ``brain'' of Chat-GPT's foundation language model, and analyze it with respect to a metric representing the notion of \emph{fairness}.
We develop a novel approach to visualize GPT's moral dimensions. We first compute a fairness metric, inspired by social psychology literature, to identify factors that typically influence fairness assessments in humans, such as legitimacy, need, and responsibility. Subsequently, we summarize the manifold's shape using a lower-dimensional simplicial complex, whose topology is derived from this metric. We color it with a heat map associated with this fairness metric, producing human-readable visualizations of the high-dimensional sentence manifold.
Our results show that sentence embeddings based on GPT-3.5 can be decomposed into two submanifolds corresponding to fair and unfair moral judgments. This indicates that GPT-based language models develop a moral dimension within their representation spaces and induce an understanding of fairness during their training process.
\end{abstract}

\newpage

\section*{INTRODUCTION}

\begin{figure}[!h]
	\centering
	\includegraphics[width=0.8\textwidth]{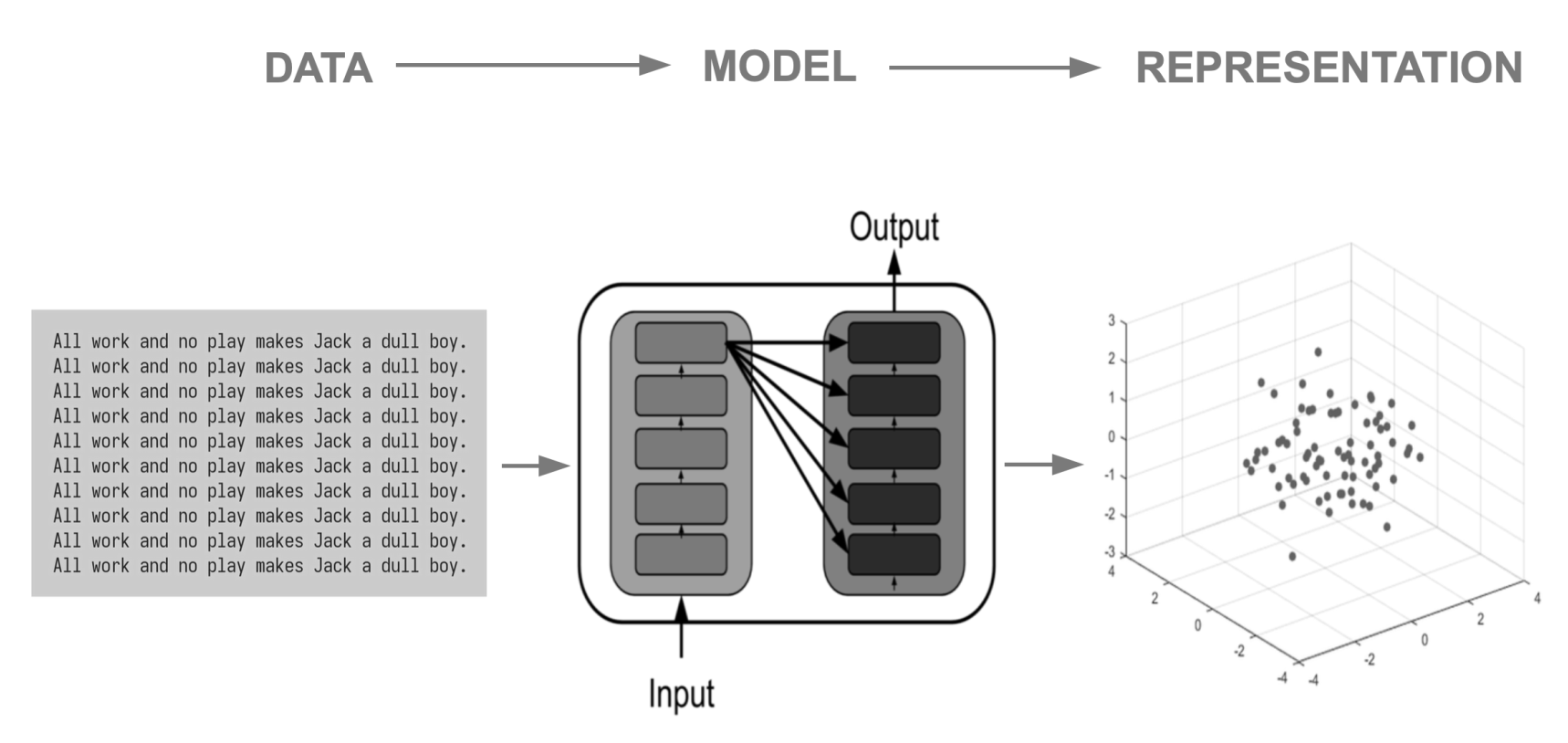}
  \caption{Large transformer language models develop their understanding of sentences by inducing informative vector space representation of linguistic units within their neural activations. The above figure summarizes this idea. A corpus of text is transformed into a point cloud of vectors by the base language model. This induces an implicit embedding of any text into a submanifold of a high dimensional ambient vector space, whose metric is learned in self-supervised fashion by backpropagation gradient descent on large amounts of text extracted from the internet. It is precisely the topology and geometry of that vector space that allow AI assistants such as Chat-GPT to perform their cognitive functions. (The figure above shows the point cloud in 3D for visualization purposes, while the manifold we analyze in this study is embedded within a much larger space $\cong \mathbb{R}^{1536}$.)}
	\label{data_model_representation}
\end{figure}

Large Language Models and Natural Language Processing systems based on them are currently at the forefront of research and applications of Artificial Intelligence.
Progress in this subfield of AI was made possible through increasingly more advanced representation methods of natural language inputs.
Initially shallow pre-training of early model layers became standard in NLP research through methods such as word2vec \cite{mikolov2013distributed}.
Subsequent progress followed trends similar to those in Computer Vision, which naturally led to pre-training of multiple layers of abstraction.
These advancements resulted in progressively deeper hierarchical language representations, such as those derived using self-attention mechanisms in transformer-based architectures \cite{vaswani2017attention}. Currently SOTA NLP systems use representations derived from pre-training of entire language models on large quantities of raw text, and often involve billions of parameters.
Those informative distributed representations of language in high dimensional vector spaces induced from large quantities of raw text are at the core of emergent abilities of Large Language Models.

Vector space representations of linguistic units within language models reflect the social propensities determined by the psychology literature, as language reflects the social values of its speakers \cite{Kennedy2021} \cite{Smith2010}.
Therefore, sentences describing fair acts will be more closely associated with sentences describing responsibility, benefit, joy, and reward than their antithesis terms of irresponsibility, harm, sadness, and punishment.
This pro-social bias can be leveraged to construct a moral compass within LLM embedding manifolds.
To this end linear spans of specially selected vectors can be used \cite{ref38} to narrow the implicit ontological associations.
Word embeddings implicitly reflect ontological knowledge \cite{Bhatia2017} \cite{Erk2012} \cite{Racharak2021} \cite{Runck2019}, such as grammatical ontologies due to the co-occurrence of specific grammatical knowledge in the co-occurrence of words \cite{ref77}.
The term "fairness," being a collection of several social ontologies, may be represented using linear combinations of hand crafted basis vectors.

The seminal work of \cite{mikolov2013distributed}, \cite{ref74}, and \cite{ref83} was foundational in providing machinery for progress in this area.
Later studies have demonstrated that language models hold implicit representations of moral values as seen in \cite{ref86} \cite{ref56} \cite{Izzidien2022}.
The authors of \cite{ref60} applied word embeddings to analyze cultural meaning and \cite{Leavy2019} used them for a study on social justice.
Moral Foundation Theory developed in \cite{ref41} was used to analyze texts, and previous studies labeled datasets with categories and applied machine learning algorithms to learn the distinctions between each category.
Similarly, the authors of \cite{Hoover2020} \cite{ref81} \cite{ref80} \cite{Araque2020} \cite{ref73} worked on predefined measures of moral language.
Prior exploration of psychological factors influencing fairness assessments includes Dictator Game and its variations, including \cite{Engel2011} \cite{ref72} \cite{Zhang2014} \cite{Branas-Garza2007} \cite{Rodrigues2015}. These studies emphasize the role of legitimacy and need in pro-social behavior.
The principal factor of responsibility is discussed with reference to several studies, such as \cite{ref94} \cite{Helson2002} \cite{ref55} \cite{Handgraaf2008}. They highlight the influence of responsibility on pro-social behavior and its dependency on cultural climate.
Additionally, contingent factors that influence responsibility perception were explored in studies by \cite{Branas-Garza2014} \cite{ref14} \cite{Chiaravutthi2019} \cite{Perera2016} for the benefit-harm gained; \cite{Gillet2009} \cite{Lejano2012} \cite{ref87} for the consideration of wider public benefit and harm; \cite{Batson1991} \cite{Edele2013} \cite{Scheres2006} \cite{Tabibnia2007} for the emotional salience of the context; \cite{Bartling2012} \cite{ref9} \cite{ref32} \cite{Henrich2001} \cite{Nesse1990} \cite{Scheres2006} \cite{ref91} for the possible consequences of rewards and punishments.
Several studies aimed to remove or diminish the role of responsibility, including \cite{Cryder2012} \cite{Hamman2010} \cite{Bartling2012}.
These studies show that when responsibility is diminished, pro-social behavior decreases.
By integrating these observations, the authors of \cite{Izzidien2022} developed a fairness metric for natural language text, that approximates human perceptions of fairness.

\begin{figure}[!h]
	\centering
	\includegraphics[width=0.8\textwidth]{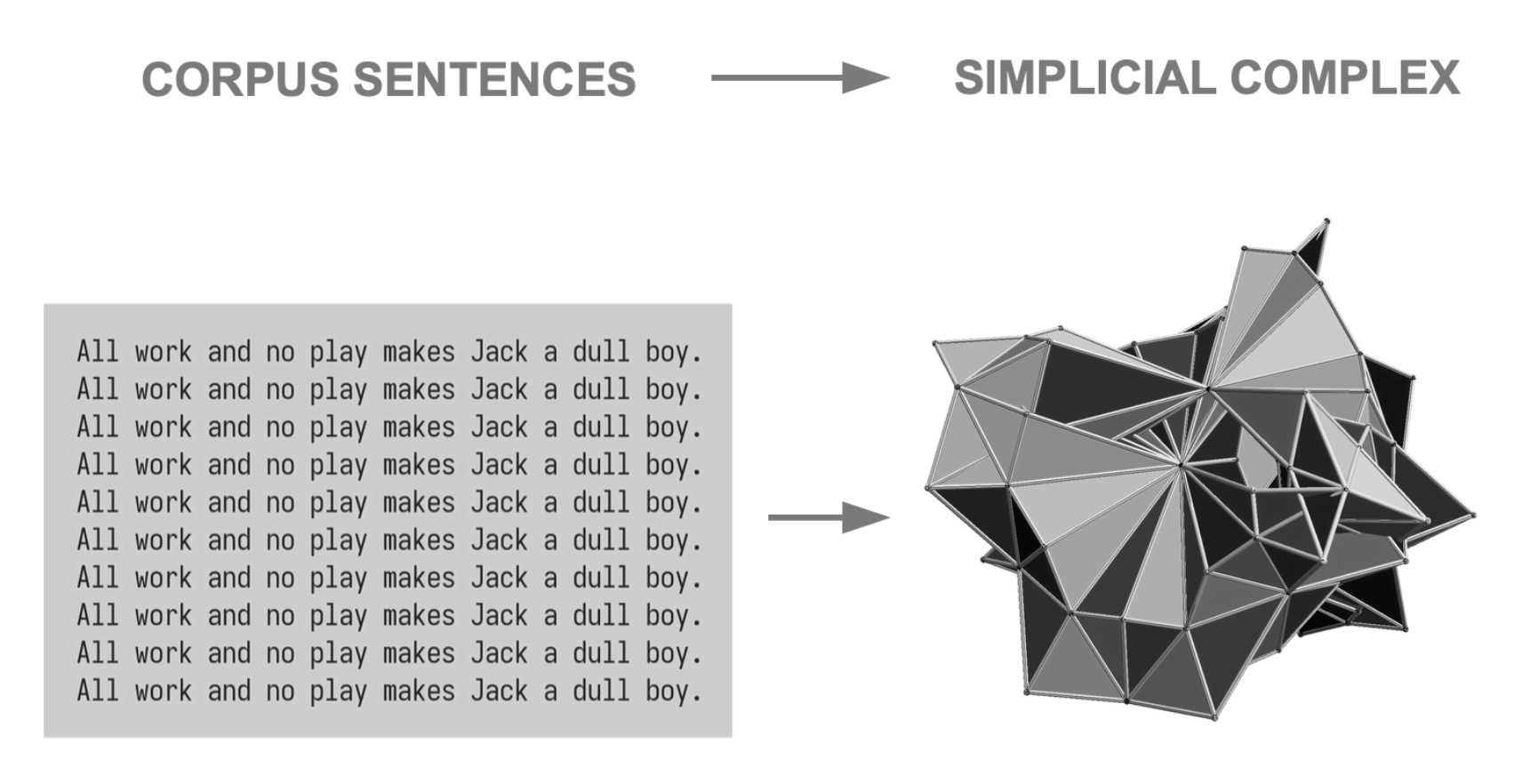}
  \caption{Vector space representations induced by LLMs such as GPT-3.5 are the result of multiple nonlinear mappings through a sequence of high dimensional ambient vector spaces. Simple dimensionality reduction techniques such as linear projections can be very misleading. We would like to visualize the distribution of sentence embeddings with respect to the fairness dimension (as defined here) while preserving some notion of shape of the original space. In order to do so, instead of simply projecting sentence embeddings onto that direction, we cluster the original data by the mapping technique described in this paper. This results in a low dimensional simplicial complex, whose shape can be thought of as a coarse grained summary of the topological features of the original high dimensional representation manifold. Here vertices correspond to clusters of sentence embeddings, while edges correspond to intersections of clusters in the projection onto the fairness direction. Higher dimensional simplices (triangles, tetrahedra, ...) would correspond to multiple clusters intersecting. In our study we study this representation only up to the graph dimension, which gives a good enough sketch of the manifold's shape to conclude separation into two submanifolds corresponding to fair versus unfair moral judgments.}
	\label{corpus_sentences_simplicial_complex}
\end{figure}

In this paper we probe the topological structure of vector space representations of sentences induced by GPT-3.5 - the foundation language model behind Chat-GPT in order to gauge its understanding of moral values.
For this purpose we introduce novel approaches from computational algebraic topology into the study of LLMs.
We compute a metric of fairness based on prior studies in computational social sciences, and use it to perform a topological summary of 1536-dimensional sentence embedding manifold with a one-dimensional simplicial complex (i.e. a graph) whose vertices correspond to clusters of sentence representations under a projection onto a "fairness" dimension, and edges to pairwise cluster intersections.
This procedure produces a sketch of the original high-dimensional manifold (which is impossible to visualize directly) with a topological object that can be visualized in human readable form.
Furthermore, the shape of the resulting visualization gives clues about the general shape of the representation space of the language model.
We also produce a heat map with colors corresponding to the degree of fairness.
This coloring allows us to inspect the manifold for separation into submanifolds corresponding to fair vs unfair moral judgments.
We observe a clear pattern of separation, which suggests the notion of fairness is encoded in the shape of the high dimensional representation manifold induced by the language model.

\section*{FAIRNESS METRIC}

\begin{figure}[!h]
	\centering
	\includegraphics[width=0.98\textwidth]{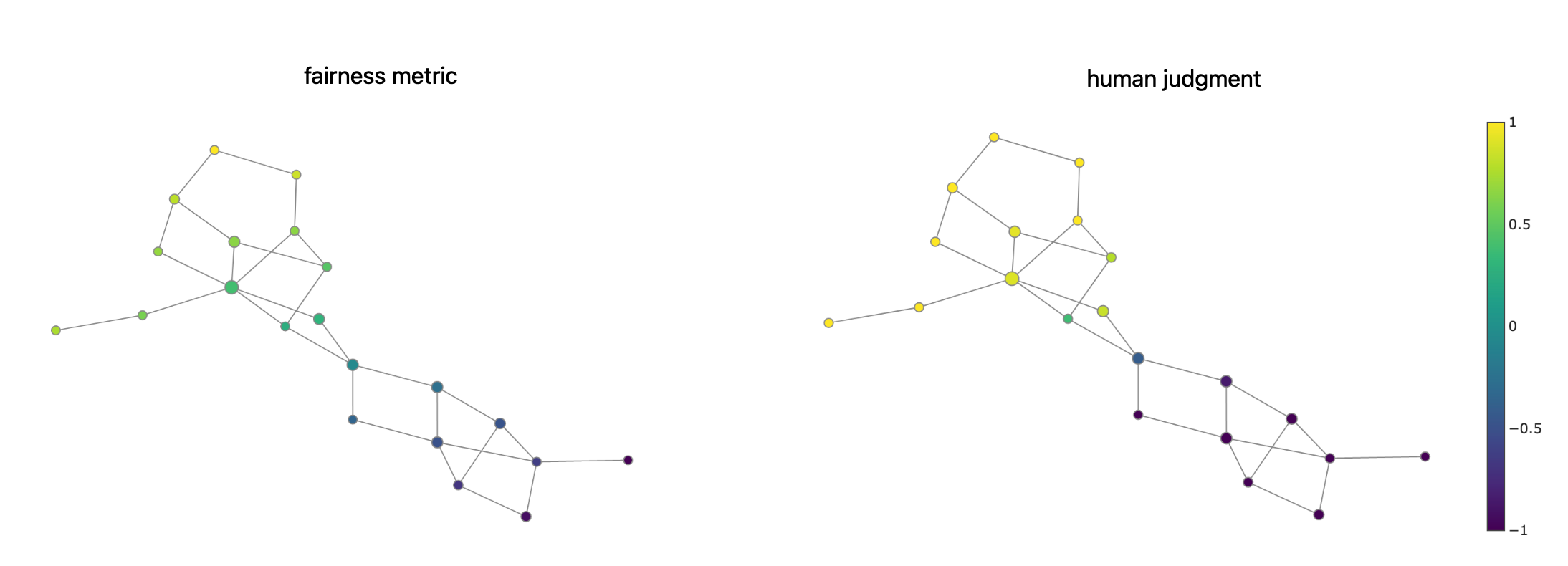}
  \caption{Topological summary of the 1536-dimensional vector space representation within GPT-3.5 of the \emph{fairness corpus} \cite{Izzidien2022} composed of equal number of fair and unfair sentences (according to human judgement). Each node is composed of multiple sentences. The size of the node corresponds to the cardinality, and the color to fairness metric (dark purple corresponding to unfair, and bright yellow corresponding to fair directions within the embedding manifold). The heat map on the left corresponds to the fairness measure based on the criterion described here (magnitude of the projection onto the fairness subspace), while the one on the right to human judgments (-1 for unfair, 1 for fair).}
	\label{fairvec}
\end{figure}

The authors of \cite{Izzidien2022} propose to harness implicit social biases in language as a metric for an explainable assessment of sentences related to fairness.
They delve into the psychology literature to identify factors that influence humans' fairness assessments. They discuss the Dictator Game and its variations, which have been used to isolate factors that predict pro-social behavior, such as legitimacy and need. They also highlight the principal factor of responsibility, which has been shown to play a critical role in pro-social behavior and fairness assessments.
We compute a fairness metric based on those ideas with respect to sentence representations induced by the foundation language model behind Chat-GPT3.5.
We subsequently use this metric to analyze the shape of high dimensional point clouds of sentence embeddings induced by the language model.

Similarly to \cite{Izzidien2022} we define the fairness dimension as a linear combination of 5 factors representing concepts of responsibility, pleasure, benefit, reward, and harm. In our computation, every factor is derived from pairs of sentences representing two polarities of each factor. The vector representations are obtained from stimulating neurons of the transformer encoder blocks within GPT3.5 by a set of several hundred examples of sentences representing fair vs unfair moral situations according to human feedback.

In particular the fairness dimension is defined to be the linear span of the vector defined by:
$$
\vec{\textbf{resV}} - \vec{\textbf{irresV}} + \vec{\textbf{joyV}} - \vec{\textbf{painV}}  + \vec{\textbf{libV}} - \vec{\textbf{priV}}   + \vec{\textbf{benV}} - \vec{\textbf{harV}}  + \vec{\textbf{appV}} - \vec{\textbf{inappV}}
$$
, where resV, irresV, joyV, painV, libV, priV, benV, harV, appV, inappV are GPT-3.5 derived vector space representations of respective concepts: \textit{"it was very respondible"}, \textit{"it was very irresponsible"}, \textit{"it was joyous"}, \textit{"it was sad"}, \textit{"it was beneficial to society"}, \textit{"it was not beneficial to society"}, \textit{"was free to and rewarded"}, \textit{"was sent to prison and punished"}, \textit{"it was beneficial"}, \textit{"it was harmful"}.

Because large GPT models are trained in autoregressive fashion on trillions of tokens of text, the metric they induce in the sentence representations encodes various features of natural language sentences, including semantics and moral judgments.
This translates to the inner product within those spaces expressing similarity across multiple aspects of natural language.
Convex combinations of sentence embeddings can be used to express a semantic gradient ranging between the chosen concepts expressed by these sentences.
Very large language models like GPT-3.5 induce surprisingly powerful sentence representations, leading to the observed emergent properties of these models.
If we consider vector space representations of sentences "it was beneficial" and "it was harmful", and compare them to a representation of a test sentence, such as "the guard helped the man" by computing cosine similarity, the result will be a score in the interval $[-1, 1]$.
The more associated the sentence is with benefit, the closer to 1 will this normalized inner product tend to be.
Whereas sentences that are more associated with harmfulness will provide an outcome closer to -1.

In figure \ref{fairvec} we summarize this by displaying side by side comparison between human judgment and the distribution of the fairness metric within GPT-3.5's text representation manifold.

\section*{METHOD OF ANALYSIS}

\begin{figure}[!h]
	\centering
	\includegraphics[width=0.8\textwidth]{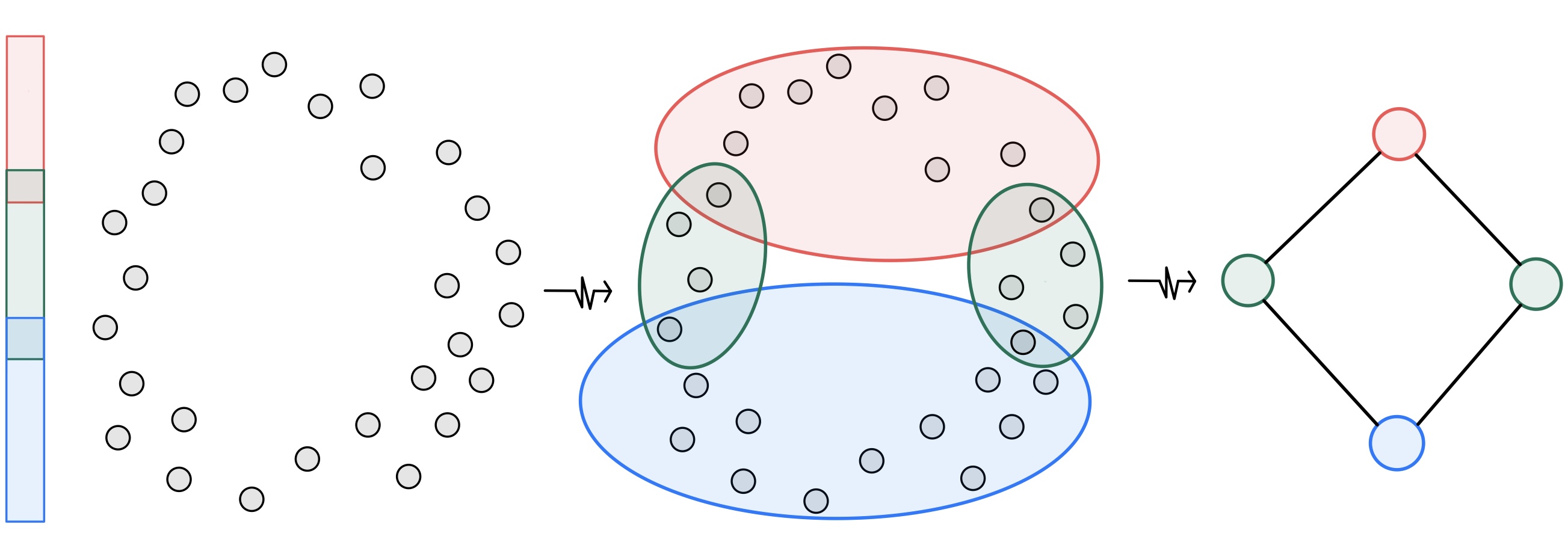}
  \caption{Inducing topological structure from a point cloud representing noisy samples from a neighborhood of a 1-dimensional submanifold of $\mathbb{R}^2$. The resulting simplicial complex is a graph with four vertices corresponding to clusters of the projection on the left, and four edges corresponding to their intersections in the original embedding space. Note that the simplicial complex obtained is homotopy equivalent to the circle $\mathbb{S}^1$ (a 1-dimensional topological manifold) and gives a reasonable summary of the shape of this original 2D point cloud of vectors.}
	\label{mapper}
\end{figure}

The technique we use to study GPT-3.5's sentence embeddings can be thought of as a topological dimensionality reduction method, where the goal is to summarize the shape of our representation space with a rough sketch in form of a low dimensional topological manifold.
This reduced representation can be thought of as a map approximating the shape of our embedding space.
Such description can be visually inspected by a human, while remaining more topologically informative than a naive projection.
Instead of studying those high dimensional vectors directly, we can map them to a different space first, define an open cover, and then cluster the original points within the preimage of each cover set.
This produces a summary of the topological features in the embedding with a simplicial complex of a chosen dimension \cite{singh2007topological}. 
In particular we generated 1-dimensional simplicial complexes (i.e. graphs) from GPT-3.5 based sentence representations by projecting them onto the fairness dimension (from the last section) and then analyzing clusters of original high dimensional sentence embeddings with respect to an open cover of that projection.
We also applied a coloring according to a heat map corresponding to the magnitude and sign of that projection.
The resulting data structure allows for a visual exploration of the shapes of these high dimensional embedding manifolds in order to identify distribution of fairness in the representation space.
Figure \ref{mapper} shows a visualization of this process for a point cloud sampled from the circle ($\mathbb{S}^2$).

The general procedure can be summarized as follows.

\begin{center}
  \setlength{\fboxsep}{1em}
  \setlength{\fboxrule}{0.1em}
  \noindent\fcolorbox{black}{lightgray}{%
    \minipage[t]{\dimexpr0.98\linewidth-2\fboxsep-2\fboxrule\relax}

      Given data points \( \mathbb{X} = \{x_1, \dots, x_n\}, x_i \in \mathbb{R}^d \), a function \( f: \mathbb{R}^d \rightarrow \mathbb{R}^m, m < d \), and a cover \( \mathcal{U} = \bigcup_{i \in \mathcal{I}} U_i \) of the image \( f(\mathbb{X}) \) (where \( \mathcal{I} \) is some index set) we construct a simplicial complex as follows:

      \begin{enumerate}
        \item For each \( U_i \in \mathcal{U} \), cluster \( f^{-1}(U_i) \) into \( k_{U_i} \) clusters \( C_{U_{i,1}}, \dots, C_{U_i,k_{U_i}} \)
        \item \( \underset{U_i \in \mathcal{U}}{\bigsqcup} \{C_{U_{i,1}}, \dots, C_{U_i,k_{U_i}}\} \) now define a cover of \( \mathbb{X} \); calculate the nerve of this cover
      \end{enumerate}

      Nerve is defined in the following way. Given a cover \( \mathcal{U} = \bigcup_{i \in \mathcal{I}} U_i \), the nerve of \( \mathcal{U} \) is the simplicial complex \( \mathcal{C}(\mathcal{U}) \) where the 0-skeleton is formed by the sets in the cover (each \( U_i \) is a vertex) and \( \sigma =[U_{j_0}, \dots, U_{j_k}] \) is a k-simplex \( \iff \bigcap\limits_{l=0}^{k} U_{l_k} \neq 0 \)
    \endminipage
  }
\end{center}

\section*{DISCUSSION OF RESULTS}

\begin{figure}[!h]
	\centering
	\includegraphics[width=0.8\textwidth]{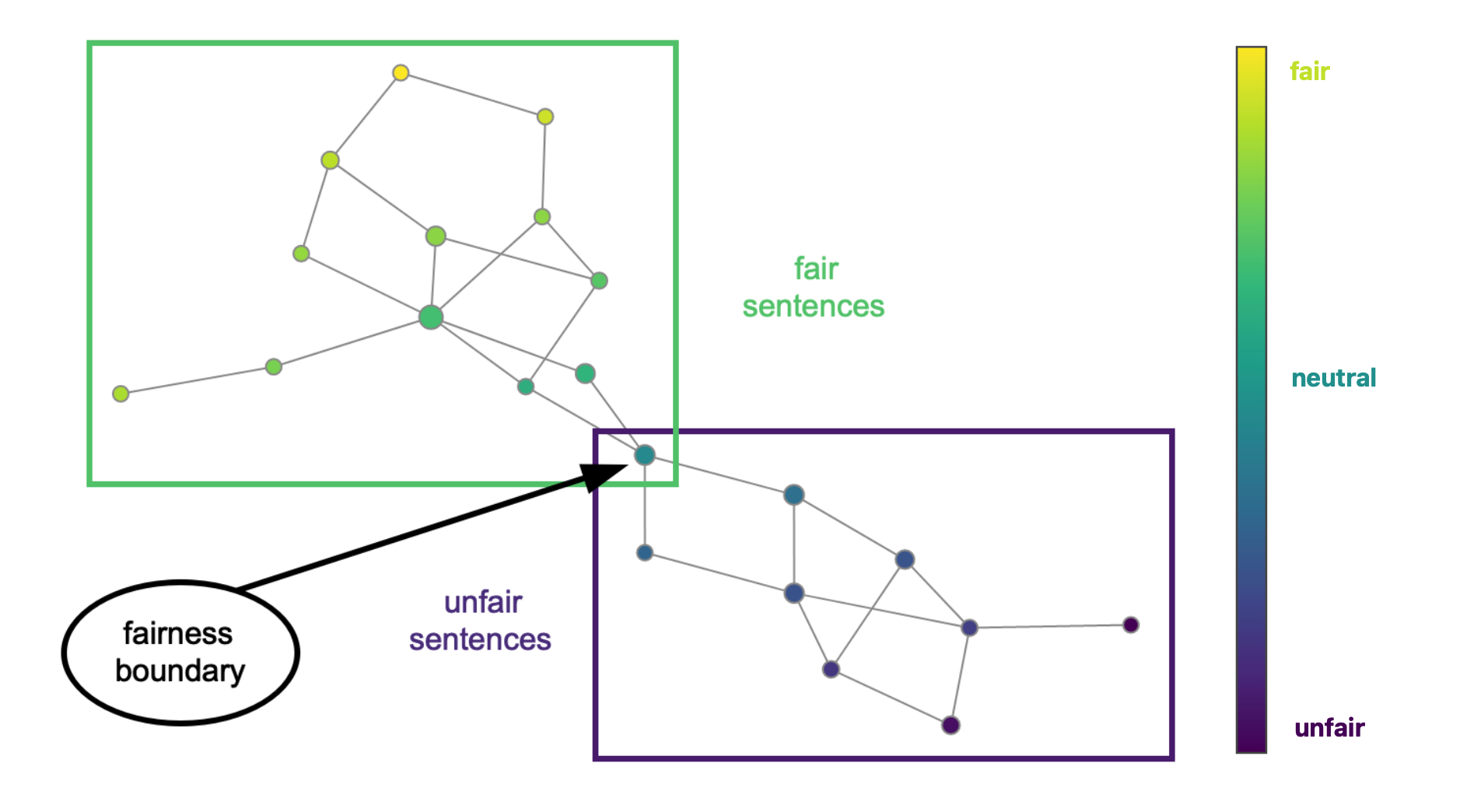}
  \caption{The fairness subspace clearly subdivides the 1536-dimensional representation space within the GPT-3.5 language model, into two submanifolds. When the language model is prompted with fair sentences, different patterns of neuronal activation are manifested within the transformer layers than when the model is prompted with morally unfair examples. This leads to the sentence representation manifold being separated into submanifolds corresponding to these two extremes. Each part is itself a topologically complex space, but the two parts are linked at a single cluster of morally ambiguous sentences. The sentences represented in the green part of the manifold are all judged to be fair, while all the sentences in the blue part represent unfair situations.}
	\label{fairness_boundary}
\end{figure}

Visualization in figure \ref{fairness_boundary} shows the results of applying our method to 1536-dimensional vector space representations of sentences obtained from GPT-3.5 based embedding model.
This is the foundation language model used by ChatGPT, and we believe its emergent abilities to judge fairness of situations are partially enabled by the topological structure of this embedding manifold.
Each node corresponds to a cluster of sentences obtained by first projecting their embeddings onto the fairness subspace defined earlier in this paper, and then grouping the original sentences based on preimages of an open cover of this projection.
This procedure gives a topological sketch of the general shape of this high dimensional sentence representation manifold.
Node sizes are defined in proportion to the number of sentences in each cluster.
Their colors are based on the heat map shown to the right, which is a gradient corresponding to the sign and magnitude of the projection.
High values on that scale correspond to fair, while low values to unfair situations (as described by the sentences).
The edges come from cluster intersections in the original high dimensional space, and thus give visual cues to how the clusters are distributed within the embedding manifold.

The two boxes contain fair and unfair sentences.
We see that each box bounds a topologically complex manifold (due to the nontrivial pattern of edges), but there are no edges connecting nodes from different boxes.
This means the entire manifold is composed of two submanifolds corresponding to fair and unfair situations.
This separation of colors suggests that there are subregions in this high dimensional representation space corresponding to activation patterns within the encoder neural network that occur only when fair sentences are processed, and separate disconnected regions corresponding to neural activations caused by unfair sentences.
Before the model is trained, this separation could not occur (other than with infinitesimal probability from random initialization), and the fact that GPT-3.5 arrives at this topology in its neural representations of language, implies it (and hence models based on it, such as ChatGPT) possess the ability to represent moral dimensions of fairness.

The behavior of these models is determined by the vector space representations of inputs they induce from text during training.
Therefore, the existence of clear patterns in the shape of those representation manifolds, such as separation into regions corresponding to moral aspects of language, can serve as a tool to intrinsically examine their abilities.
This is in contrast to currently used mainstream approaches, which are extrinsic and behavioral in nature.
In these contemporary approaches the abilities of the model are examined with techniques based on prompting and analysing the outputs, or by proxy of performance on downstream tasks.
We believe the approach presented here is novel in the field of AI, and NLP in particular.
It is more analogous to performing studies on the brain with something like fMRI or a Neuralink device, and analyzing topological patterns in neural communication in order to study the subject, instead of making judgment based on the subject's apparent behavior, or reports from introspection.
We believe this alternative way of looking at language models could benefit alignment research and AI safety, and we hope this paper serves as an inspiration for the community to develop more tools of this kind.

\bibliographystyle{unsrt}
\bibliography{main}

\begin{thebibliography}{10}

\bibitem{mikolov2013distributed}
Tomas Mikolov, Ilya Sutskever, Kai Chen, Greg~S Corrado, and Jeff Dean.
\newblock Distributed representations of words and phrases and their
  compositionality.
\newblock In {\em Advances in neural information processing systems}, pages
  3111--3119, 2013.

\bibitem{vaswani2017attention}
Ashish Vaswani, Noam Shazeer, Niki Parmar, Jakob Uszkoreit, Llion Jones,
  Aidan~N Gomez, {\L}ukasz Kaiser, and Illia Polosukhin.
\newblock Attention is all you need.
\newblock In {\em Advances in neural information processing systems}, pages
  5998--6008, 2017.

\bibitem{Kennedy2021}
B.~Kennedy, M.~Atari, A.~Mostafazadeh~Davani, J.~Hoover, A.~Omrani, J.~Graham,
  and M.~Dehghani.
\newblock Moral concerns are differentially observable in language.
\newblock {\em Cognition}, 212, 2021.

\bibitem{Smith2010}
E.~A. Smith.
\newblock Communication and collective action: Language and the evolution of
  human cooperation.
\newblock {\em Evolution and Human Behavior}, 31, 2010.

\bibitem{ref38}
Foley, d., {\&} kalita, j. (2016). integrating wordnet for multiple sense
  embeddings in vector semantics. proceedings of the 13th international
  conference on natural language processing, 2--9.

\bibitem{Bhatia2017}
S.~Bhatia.
\newblock Associative judgment and vector space semantics.
\newblock {\em Psychological Review}, 124, 2017.

\bibitem{Erk2012}
K.~Erk.
\newblock Vector space models of word meaning and phrase meaning: A survey.
\newblock {\em Language and Linguistics Compass}, 6, 2012.

\bibitem{Racharak2021}
T.~Racharak.
\newblock On approximation of concept similarity measure in description logic
  elh with pre-trained word embedding.
\newblock {\em IEEE Access: Practical Innovations, Open Solutions}, 9, 2021.

\bibitem{Runck2019}
B.~C. Runck, S.~Manson, E.~Shook, M.~Gini, and N.~Jordan.
\newblock Using word embeddings to generate data-driven human agent
  decision-making from natural language.
\newblock {\em GeoInformatica}, 23, 2019.

\bibitem{ref77}
Qian, p., qiu, x., {\&} huang, x. (2016). investigating language universal and
  specific properties in word embeddings. proceedings of the 54th annual
  meeting of the association for computational linguistics (volume 1: Long
  papers), 1478--1488. https://doi.org/10.18653/v1/p16-1140.

\bibitem{ref74}
Pennington, j., socher, r., {\&} manning, c. d. (2014). glove: Global vectors
  for word representation. proceedings of the 2014 conference on empirical
  methods in natural language processing (emnlp), 1532--1543.

\bibitem{ref83}
Rong, x. (2014). word2vec parameter learning explained. arxiv prepr,
  arxiv14112738.

\bibitem{ref86}
Schramowski, p., turan, c., jentzsch, s., rothkopf, c., {\&} kersting, k.
  (2019). bert has a moral compass: Improvements of ethical and moral values of
  machines. arxiv preprint arxiv:1912.05238.

\bibitem{ref56}
Jentzsch, s., schramowski, p., rothkopf, c., {\&} kersting, k. (2019).
  semantics derived automatically from language corpora contain human-like
  moral choices. proceedings of the 2019 aaai/acm conference on ai, ethics, and
  society, 37--44.

\bibitem{Izzidien2022}
A.~Izzidien.
\newblock Word vector embeddings hold social ontological relations capable of
  reflecting meaningful fairness assessments.
\newblock {\em AI {\&} Society}, 37, 2022.

\bibitem{ref60}
Kozlowski, a. c., taddy, m., {\&} evans, j. a. (2019). the geometry of culture:
  Analyzing the meanings of class through word embeddings. the american
  sociological review, 84(5), 905--949.

\bibitem{Leavy2019}
S.~Leavy, M.~T. Keane, and E.~Pine.
\newblock Patterns in language: Text analysis of government reports on the
  irish industrial school system with word embedding.
\newblock {\em Digital Scholarship in the Humanities}, 34, 2019.

\bibitem{ref41}
Graham, j., haidt, j., koleva, s., motyl, m., iyer, r., wojcik, s. p., {\&}
  ditto, p. h. (2013). moral foundations theory: The pragmatic validity of
  moral pluralism. advances in experimental social psychology (47 vol., pp.
  55--130). elsevier.

\bibitem{Hoover2020}
J.~Hoover, G.~Portillo-Wightman, L.~Yeh, S.~Havaldar, A.~M. Davani, Y.~Lin,
  B.~Kennedy, M.~Atari, Z.~Kamel, and M.~Mendlen.
\newblock Moral foundations twitter corpus: A collection of 35k tweets
  annotated for moral sentiment.
\newblock {\em Social Psychological and Personality Science}, 11, 2020.

\bibitem{ref81}
Rezapour, r., shah, s. h., {\&} diesner, j. (2019). enhancing the measurement
  of social effects by capturing morality. proceedings of the tenth workshop on
  computational approaches to subjectivity, sentiment and social media
  analysis, 35--45.

\bibitem{ref80}
Rezapour, r., dinh, l., {\&} diesner, j. (2021). incorporating the measurement
  of moral foundations theory into analyzing stances on controversial topics.
  proceedings of the 32nd acm conference on hypertext and social media,
  177--188.

\bibitem{Araque2020}
O.~Araque, L.~Gatti, and K.~Kalimeri.
\newblock Moralstrength: Exploiting a moral lexicon and embedding similarity
  for moral foundations prediction.
\newblock {\em Knowledge-Based Systems}, 191, 2020.

\bibitem{ref73}
Pennebaker, j. w., francis, m. e., {\&} booth, r. j. (2001). linguistic inquiry
  and word count: Liwc 2001. mahway: Lawrence erlbaum associates, 71(2001),
  2001.

\bibitem{Engel2011}
C.~Engel.
\newblock Dictator games: A meta study.
\newblock {\em Experimental Economics}, 14, 2011.

\bibitem{ref72}
Ortman, a., {\&} zhang, l. (2013). exploring the meaning of significance in
  experimental economics (no. 2013--32; discussion papers). school of
  economics, the university of new south wales.
  https://ideas.repec.org/p/swe/wpaper/2013-32.html.  retrieved december 21,
  2021.

\bibitem{Zhang2014}
L.~Zhang and A.~Ortmann.
\newblock The effects of the take-option in dictator-game experiments: A
  comment on engel's (2011) meta-study.
\newblock {\em Experimental Economics}, 17, 2014.

\bibitem{Branas-Garza2007}
P.~Bra{\~{n}}as-Garza.
\newblock Promoting helping behavior with framing in dictator games.
\newblock {\em Journal of Economic Psychology}, 28, 2007.

\bibitem{Rodrigues2015}
J.~Rodrigues, N.~Ulrich, and J.~Hewig.
\newblock A neural signature of fairness in altruism: A game of theta?
\newblock {\em Social Neuroscience}, 10, 2015.

\bibitem{ref94}
Tisserand, j. c., cochard, f., {\&} le gallo, j. (2015). altruistic or
  strategic considerations: A meta-analysis on the ultimatum and dictator
  games. besan{\c{c}}on: Crese, universit{\'e} de franche-comt{\'e}.

\bibitem{Helson2002}
R.~Helson, C.~Jones, and V.~S.~Y. Kwan.
\newblock Personality change over 40 years of adulthood: Hierarchical linear
  modeling analyses of two longitudinal samples.
\newblock {\em Journal of Personality and Social Psychology}, 83, 2002.

\bibitem{ref55}
Jensen-campbell, l. a., knack, j. m., {\&} rex-lear, m. (2009). personality and
  social relations. the cambridge handbook of personality
  psychology. core/books/cambridge-handbook-of-personality-psychology/personality-and-social-relations/342aedb44a6cd3e82ab89e2f353b01d0. 
  retrieved december 21, 2021.

\bibitem{Handgraaf2008}
M.~J. Handgraaf, E.~Dijk, R.~C. Vermunt, H.~A. Wilke, and C.~K. Dreu.
\newblock Less power or powerless? egocentric empathy gaps and the irony of
  having little versus no power in social decision making.
\newblock {\em Journal of Personality and Social Psychology}, 95, 2008.

\bibitem{Branas-Garza2014}
P.~Bra{\~{n}}as-Garza, A.~M. Esp{\'i}n, F.~Exadaktylos, and B.~Herrmann.
\newblock Fair and unfair punishers coexist in the ultimatum game.
\newblock {\em Scientific Reports}, 4, 2014.

\bibitem{ref14}
Kopec, m., {\&} bruner, j. (2022). no harm done? an experimental approach to
  the nonidentity problem. journal of the american philosophical association,
  8(1), 169--189. https://doi.org/10.1017/apa.2021.1.

\bibitem{Chiaravutthi2019}
Y.~Chiaravutthi.
\newblock Ethical orientation versus short-term ethics training: Effects on
  ethical behavior in the prisoner's dilemma game and dictator game
  experiments.
\newblock {\em DLSU Business {\&} Economics Review}, 29, 2019.

\bibitem{Perera2016}
P.~Perera, E.~Canic, and E.~A. Ludvig.
\newblock Cruel to be kind but not cruel for cash: Harm aversion in the
  dictator game.
\newblock {\em Psychonomic Bulletin {\&} Review}, 23, 2016.

\bibitem{Gillet2009}
J.~Gillet, A.~Schram, and J.~Sonnemans.
\newblock The tragedy of the commons revisited: The importance of group
  decision-making.
\newblock {\em Journal of Public Economics}, 93, 2009.

\bibitem{Lejano2012}
R.~P. Lejano and H.~Ingram.
\newblock Modeling the commons as a game with vector payoffs.
\newblock {\em Journal of Theoretical Politics}, 24, 2012.

\bibitem{ref87}
Sigmund, k., hauert, c., {\&} nowak, m. a. (2001). reward and punishment.
  proceedings of the national academy of sciences, 98(19), 10757--10762.
  https://doi.org/10.1073/pnas.161155698.

\bibitem{Batson1991}
C.~D. Batson, J.~G. Batson, J.~K. Slingsby, K.~L. Harrell, H.~M. Peekna, and
  R.~M. Todd.
\newblock Empathic joy and the empathy-altruism hypothesis.
\newblock {\em Journal of Personality and Social Psychology}, 61, 1991.

\bibitem{Edele2013}
A.~Edele, I.~Dziobek, and M.~Keller.
\newblock Explaining altruistic sharing in the dictator game: The role of
  affective empathy, cognitive empathy, and justice sensitivity.
\newblock {\em Learning and Individual Differences}, 24, 2013.

\bibitem{Scheres2006}
A.~Scheres and A.~G. Sanfey.
\newblock Individual differences in decision making: Drive and reward
  responsiveness affect strategic bargaining in economic games.
\newblock {\em Behavioral and Brain Functions}, 2, 2006.

\bibitem{Tabibnia2007}
G.~Tabibnia and M.~D. Lieberman.
\newblock Fairness and cooperation are rewarding: Evidence from social
  cognitive neuroscience.
\newblock {\em Annals of the New York Academy of Sciences}, 1118, 2007.

\bibitem{Bartling2012}
B.~Bartling and U.~Fischbacher.
\newblock Shifting the blame: On delegation and responsibility.
\newblock {\em The Review of Economic Studies}, 79, 2012.

\bibitem{ref9}
Boyd, r., gintis, h., bowles, s., {\&} richerson, p. j. (2003). the evolution
  of altruistic punishment. proceedings of the national academy of sciences,
  100(6), 3531--3535. https://doi.org/10.1073/pnas.0630443100.

\bibitem{ref32}
El mouden, c., burton-chellew, m., gardner, a., {\&} west, a. (2012). what do
  humans maximize? in s. okasha {\&} k. binmore (eds.), evolution and
  rationality: Decisions, co-operation and strategic behaviour. cambridge:
  Cambridge university press. https://doi.org/10.1017/cbo9780511792601.

\bibitem{Henrich2001}
J.~Henrich, R.~Boyd, S.~Bowles, C.~Camerer, E.~Fehr, H.~Gintis, and
  R.~McElreath.
\newblock Cooperation, reciprocity and punishment in fifteen small-scale
  societies.
\newblock {\em American Economic Review}, 91, 2001.

\bibitem{Nesse1990}
R.~M. Nesse.
\newblock Evolutionary explanations of emotions.
\newblock {\em Human Nature}, 1, 1990.

\bibitem{ref91}
Strang, s., {\&} park, s. q. (2017). human cooperation and its underlying
  mechanisms. in m. w{\"o}hr {\&} s. krach (eds.), social behavior from rodents
  to humans: Neural foundations and clinical implications (pp.223--239).
  springer international publishing.
  https://doi.org/10.1007/7854{\_}2016{\_}445.

\bibitem{Cryder2012}
C.~E. Cryder and G.~Loewenstein.
\newblock Responsibility: The tie that binds.
\newblock {\em Journal of Experimental Social Psychology}, 48, 2012.

\bibitem{Hamman2010}
J.~R. Hamman, G.~Loewenstein, and R.~A. Weber.
\newblock Self-interest through delegation: An additional rationale for the
  principal-agent relationship.
\newblock {\em American Economic Review}, 100, 2010.

\bibitem{singh2007topological}
Gurjeet Singh, Facundo M{\'e}moli, Gunnar~E Carlsson, et~al.
\newblock Topological methods for the analysis of high dimensional data sets
  and 3d object recognition.
\newblock {\em PBG@ Eurographics}, 2, 2007.

\end{thebibliography}

\end{document}